\documentclass[10pt]{article}

\usepackage[vmargin=1in,hmargin=1.5in]{geometry}
\usepackage[pdftex]{lscape}

\usepackage[numbers,sort&compress]{natbib} 
\usepackage{hyperref}
\usepackage[all]{hypcap} 

\hypersetup{
  bookmarks=true,
  pdfborder={0 0 0},
  colorlinks=true,
  linkcolor=black,    
  citecolor=black,    
  filecolor=black,    
  urlcolor=black      
}

\usepackage[pdftex]{graphicx}
\DeclareGraphicsExtensions{.pdf,.png}

\usepackage[cmex10]{amsmath}
\usepackage{amssymb}

\usepackage{units}

\usepackage{array}

\usepackage[caption=false,font=small]{subfig}

\usepackage{url}

\newcommand{\ie}{\textit{i.e.,}\ }
\newcommand{\eg}{\textit{e.g.,}\ } 

\newcommand{\1}[1]{\mathbf{1}\big( #1 \big)}

\newcommand{\figref}[1]{\hyperref[#1]{Fig.~\ref*{#1}}}

\begin{document}
\title{Experimental Characterization of a Bearing-only Sensor for Use With the PHD Filter%
\thanks{This work will also appear in \cite{Dames2015b}.}}

\author{Philip~Dames~and~Vijay~Kumar
\thanks{Philip Dames and Vijay Kumar are with the Department
of Mechanical Engineering and Applied Mechanics, University of Pennsylvania, Philadelphia,
PA, 19104 USA e-mail: \texttt{\{pdames,kumar\}@seas.upenn.edu}.}}

\date{}

\maketitle

\begin{abstract}
This report outlines the procedure and results of an experiment to characterize a bearing-only sensor for use with PHD filter.
The resulting detection, measurement, and clutter models are used for hardware and simulated experiments with a team of mobile robots autonomously seeking an unknown number of objects of interest in an office environment.
\end{abstract}

\section{Introduction}
\label{sec:intro}
We are interested in applications such as environmental monitoring, search and rescue, precision agriculture, and landmark registration, in which teams of mobile robots can be used to explore an environment to search for objects of interest.
In each of these scenarios, the number of objects is not known a priori, and can potentially be very large.
Such situations are not a trivial extension of the single object case.
In general, the sensor can experience false negatives (\ie missing objects that are actually there), false positives (\ie seeing objects that are not actually there), and even when a true detection is made it is potentially a very noisy estimate.
Another challenge faced in multi-target tracking is the task of data association, or matching measurements to objects.
The number of such associations grows combinatorially with the number of objects and measurements, and it is often not possible to uniquely identify individual objects.

We track the distribution of multiple objects using the Probability Hypothesis Density (PHD) filter.
The PHD models the density of objects in the environment \cite{Mahler2007Book} and the PHD filter, from \citet{Mahler2003}, simultaneously estimates the number of objects and their locations in a computationally tractable manner.
We use the sensor models from this report in hardware and simulated experiments where a team of mobile robots autonomously detects and localizes an unknown number of objects using an information-based, receding horizon control law \cite{Dames2015a}.
See \citet{Dames2015a} for further motivation of the problem area and detail on the problem formulation and the estimation and control algorithms.

Each robot is equipped with a sensor that is able to detect objects within its field of view.
The probability of a sensor with pose $q$ detecting a target at $x$ is given by $p_d(x; q)$ and is identically zero outside of the field of view.
Note the dependence on the sensor's position, denoted by the argument $q$.
If a target at $x$ is detected by a robot at $q$ then it returns a measurement $z \sim g(z \mid x; \, q)$.
The false positive, or clutter, model consists of a PHD $c(z; q)$ describing the likelihood of clutter measurements in measurement space and the expected clutter cardinality.
Note that in general the clutter may depend on environmental factors.
The remainder of this report details the experimental system, descriptions of the three sensor models, and the resulting best-fit models to the collected data.

\section{Experimental System}
\label{sec:experimental_system}
\begin{figure}[tbp]
\centering
\includegraphics[width=0.95\columnwidth]{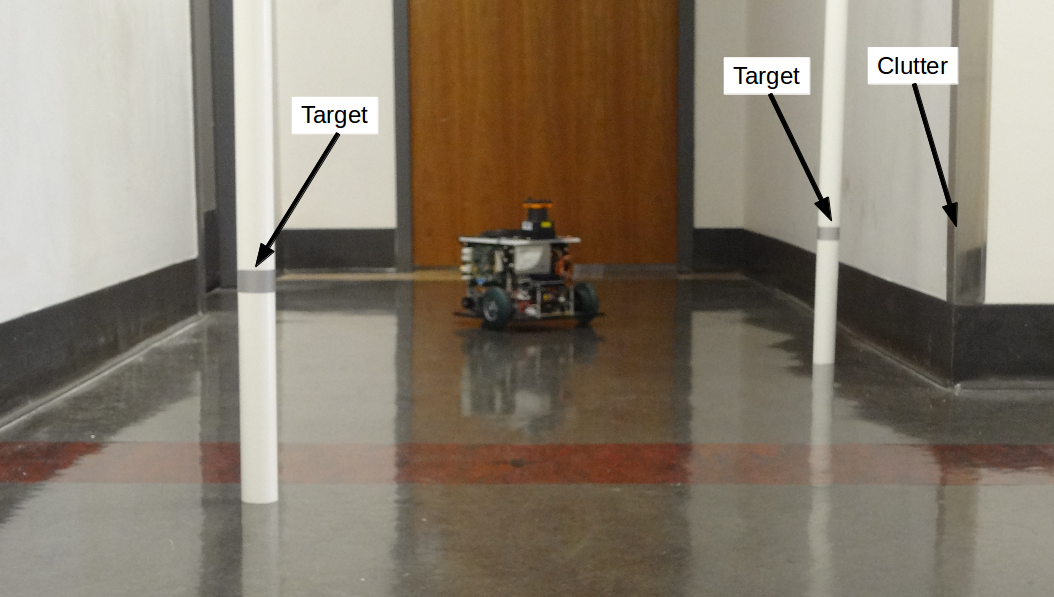}
\caption{A Scarab robot with two objects in the experimental environment.}
\label{fig:scarab_with_targets}
\end{figure}

\begin{figure}[tbp]
\centering
\includegraphics[width=0.75\columnwidth]{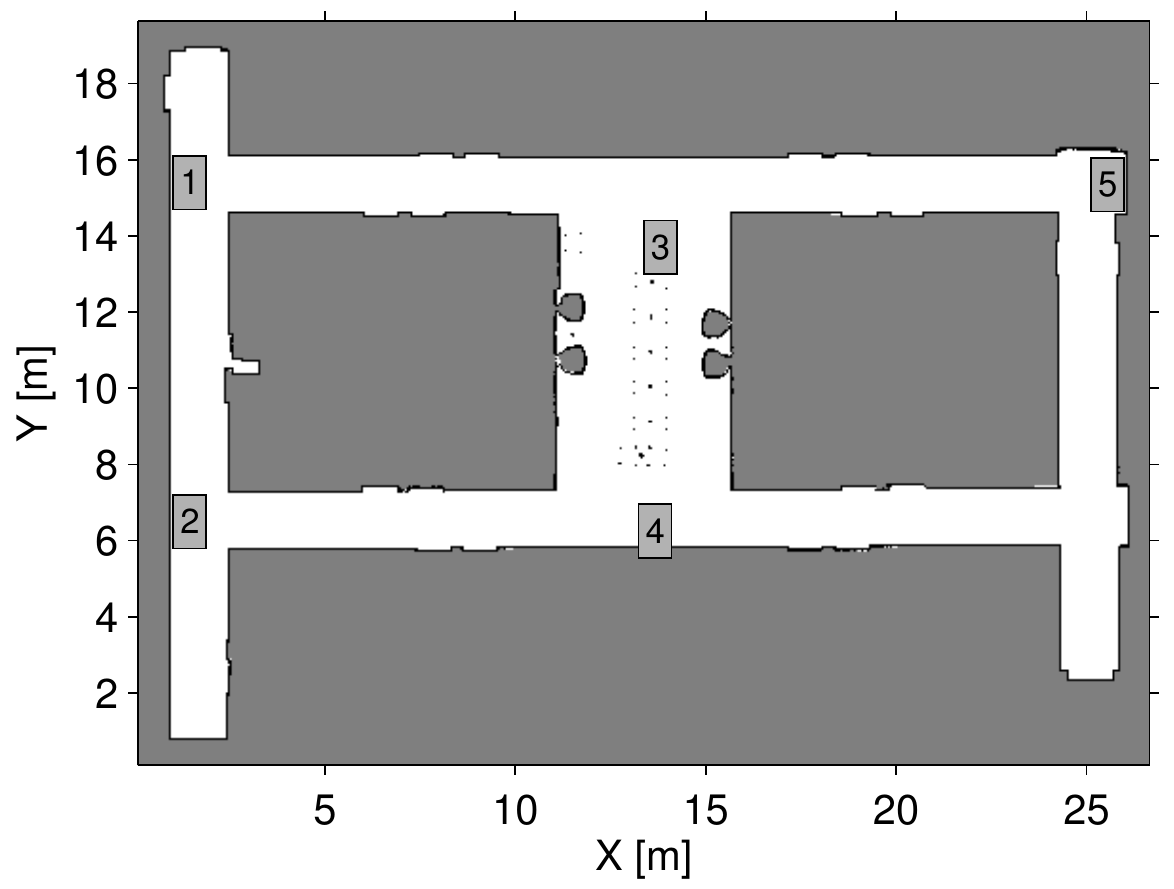}
\caption{A floorplan of the Levine environment used in the hardware experiments.  Different starting locations for the robots are labeled in the map.}
\label{fig:levine_floorplan}
\end{figure}

We conduct experiments using a small team of ground robots (Scarabs), pictured in \figref{fig:scarab_with_targets}.
The Scarabs are differential drive robots with an onboard computer with an Intel i5 processor and 4GB of RAM, running Ubuntu 12.04.
They are equipped with a Hokuyo UTM-30LX laser scanner, used for self-localization and for target detection.
The robots communicate with a central computer, a laptop with an Intel i7 processor and 16GB of RAM running ROS on Ubuntu 12.04, via an 802.11n network.
The team explores in an indoor hallway, shown in \figref{fig:levine_floorplan}, seeking the reflective objects pictured with the robot in \figref{fig:scarab_with_targets}.

We use a simple sensor, by converting the Hokuyo into a bearing-only sensor.
This may be thought of as a proxy to a camera, but it avoids common problems with visual sensors such as variable lighting conditions and distortions.
The objects are \unit[1.625]{in} outer diameter PVC pipes with attached 3M 7610 reflective tape.
The tape provides high intensity returns to the laser scanner, allowing us to pick out objects from the background environment.
However, there is no way to uniquely identify individual objects, making this the ideal setting to use the PHD filter.

\begin{landscape}
\begin{figure}[tbp]
\vspace*{-13mm}
\centering
\subfloat[XY laser scan]{
    \includegraphics[width=0.41\columnwidth]{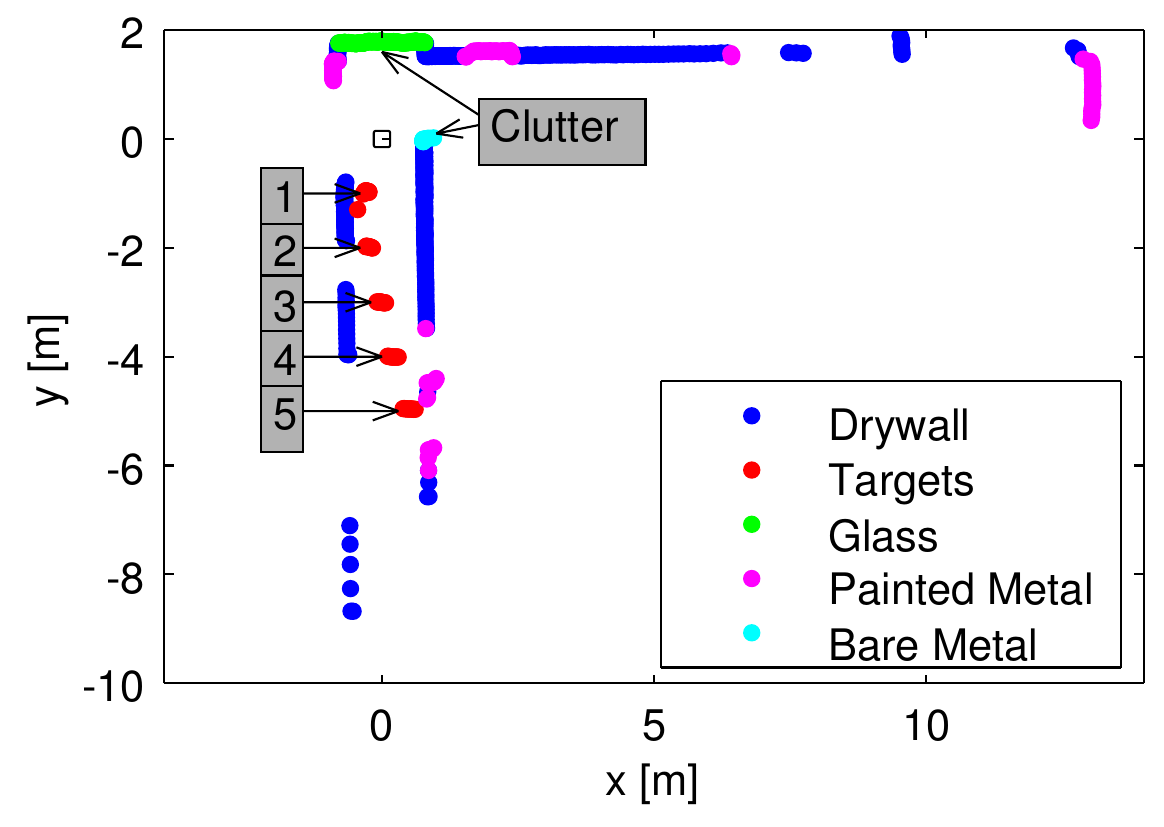}
	\label{fig:labeled_scan}
} \hspace*{2cm}
\subfloat[Intensity plot and measurement set]{
	\includegraphics[width=0.37\columnwidth]{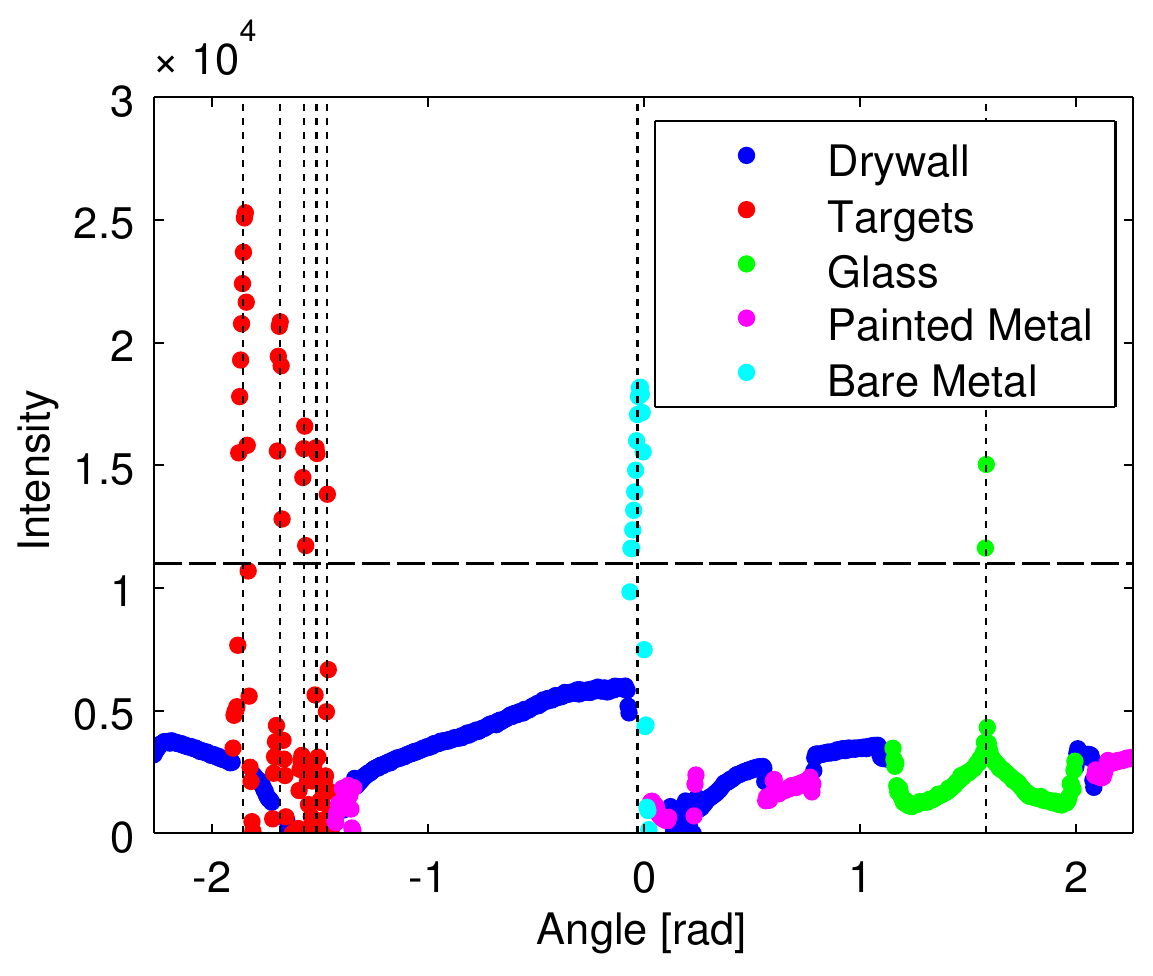}
	\label{fig:labeled_intensity}
} \\
\subfloat[Targets]{
	\includegraphics[width=0.27\columnwidth]{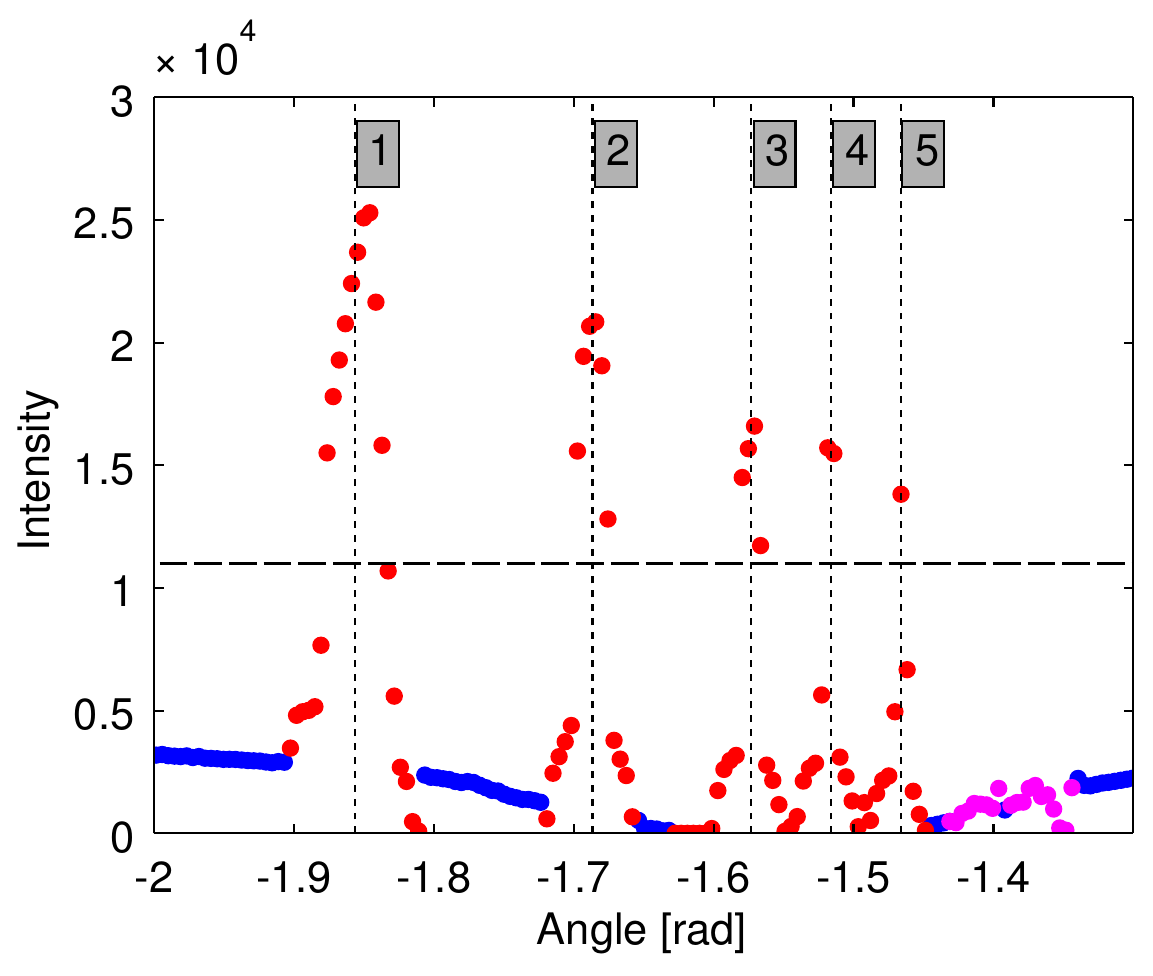}
	\label{fig:labeled_intensity_targets}
} \hfill
\subfloat[Glass]{
	\includegraphics[width=0.27\columnwidth]{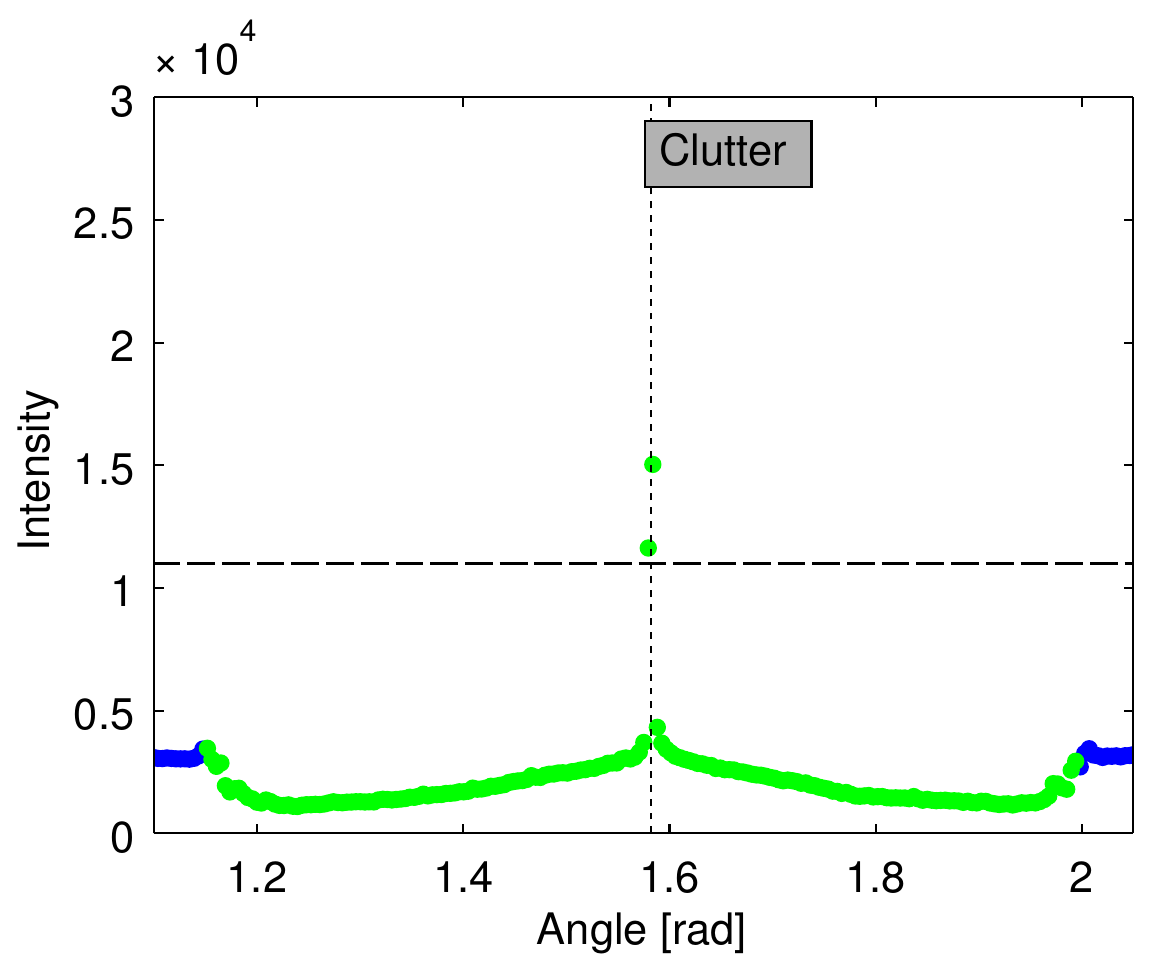}
	\label{fig:labeled_intensity_glass}
} \hfill
\subfloat[Metal]{
	\includegraphics[width=0.27\columnwidth]{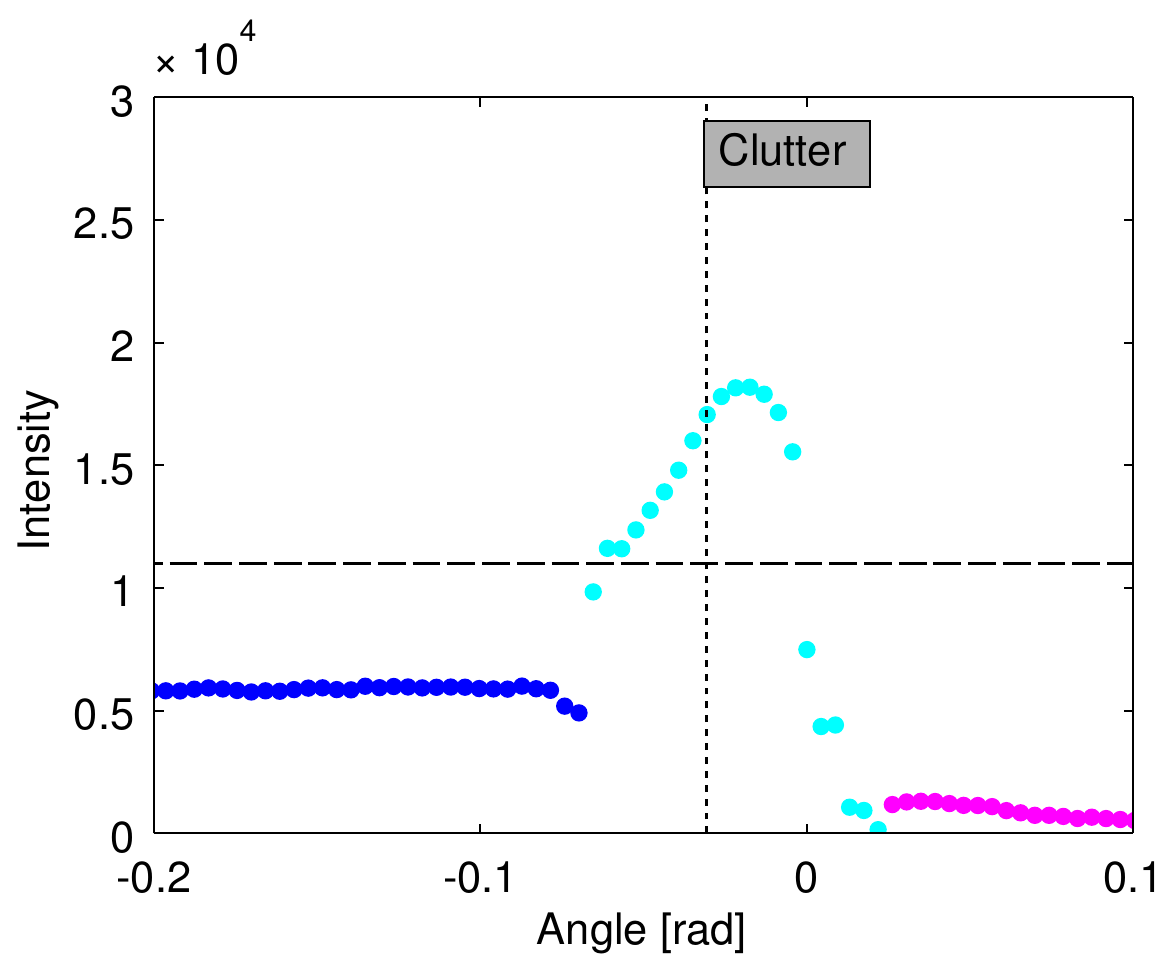}
	\label{fig:labeled_intensity_metal}
} \hfill
\caption{An example laser scan from the Levine environment. 
(a) Shows the XY scan labeled according to the building material.  Objects 1--5 are placed \unit[1--5]{m} from the robot and the sources of clutter measurements are also labeled. 
(b) Shows the corresponding intensity plot with material labels and the resulting measurement set.  
(c)-(e) Show insets of specific objects of interest within the scan and the resulting measurements.}
\label{fig:example_scan}
\end{figure}
\end{landscape}

The hallway features a variety of building materials such as drywall, wooden doors, painted metal (door frames), glass (office windows), and bare metal (chair legs, access panels, and drywall corner protectors, like that in the right side of \figref{fig:scarab_with_targets}).
The reflective properties of the environment vary according to the material and the angle of incidence of the laser.
The intensity of bare metal and glass surfaces at low angles of incidence is similar to that of the reflective tape.
\figref{fig:example_scan} shows an example laser scan from the environment, with the robot at $(0, 0)$ and oriented along the x-axis.
The objects show up as clear, high intensity sections in the laser scan, though the intensity decreases with distance to the robot (objects 1--5 are placed \unit[1--5]{m} from the robot, respectively).

Motivated by this, we select a threshold on the laser intensity of 11000 (shown as a black dotted line in the plots) to be able to reliably detect objects within a \unit[5]{m} range of the robot.
At low angles of incidence, bare metal and glass have laser returns of similarly intensity to the true objects, creating clutter measurements.
Note that glass has an extremely narrow band of angles of incidence ($\approx 0.02^\circ$) that create high intensity returns, while the metal has a slightly wider range ($\approx 2.5^\circ$).
If we were to eliminate the clutter detections, then the effective sensing range of the robots would only be less than \unit[2]{m}, significantly decreasing the utility of the system.

To turn a laser scan into a set of bearing measurements, we first prune the points based on the laser intensity threshold, retaining only those with sufficiently high intensity returns.
The points are clustered spatially using the range and bearing information, with each cluster having a maximum diameter $d_t$.
The range data is otherwise discarded.
The bearings to each of the resulting clusters form a measurement set $Z$.

\begin{figure}[tbp]
\centering
\includegraphics[width=0.7\columnwidth]{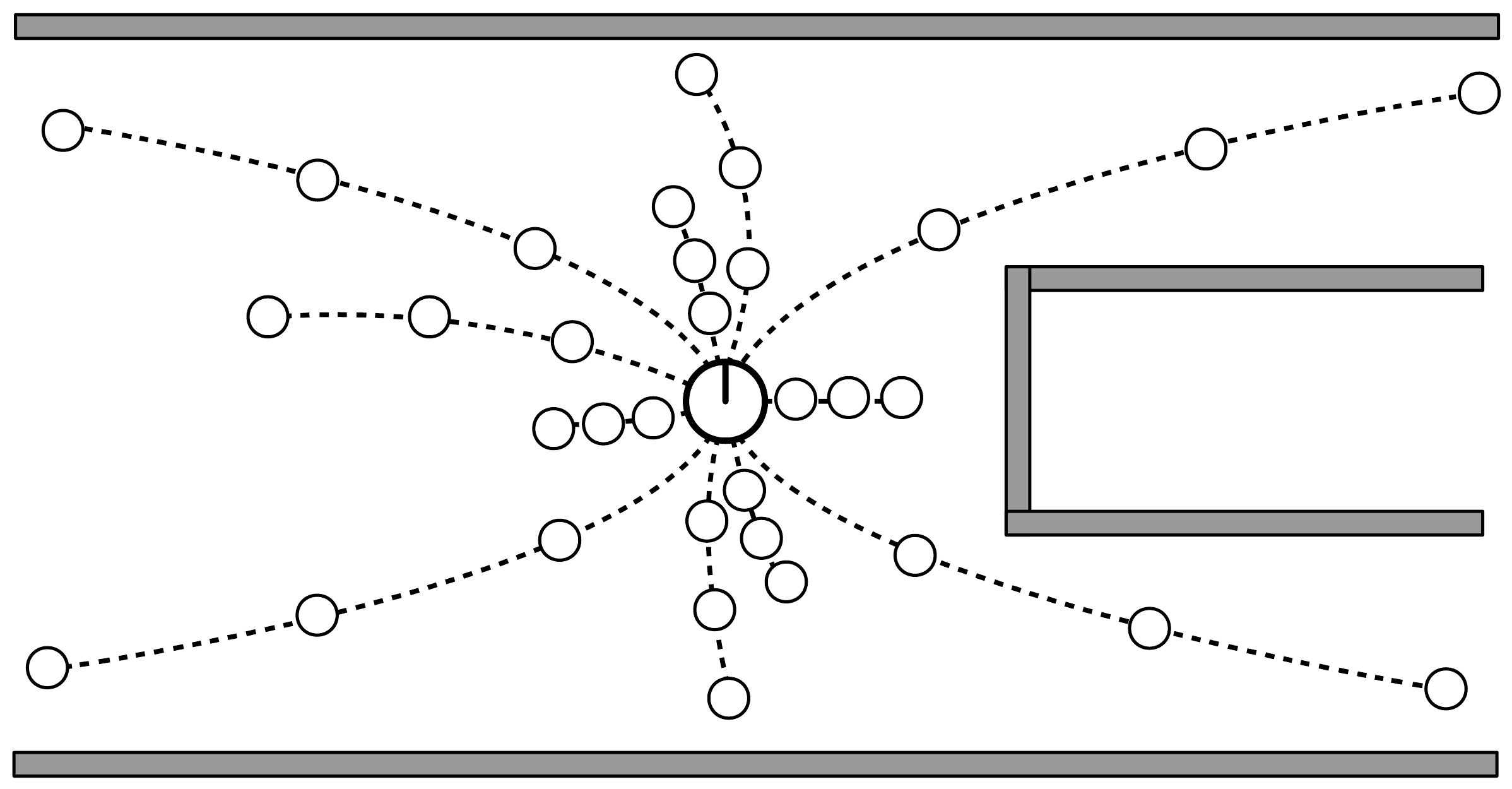}
\caption{Example action set with a horizon of $T = 3$ steps and three length scales.  Each action is a sequence of $T$ poses at which the robot will take a measurement, denoted by the hollow circles.}
\label{fig:trajectories}
\end{figure}

A team of three robots drove around the environment for \unit[20]{min} collecting measurements of 15 targets at known positions.
To move around the environment, the robots generated actions over length scales ranging from \unit[1]{m} to \unit[20]{m}, as shown in \figref{fig:trajectories}, and randomly selected actions from the candidate set.
This allowed the robots to cover the environment much more effectively than a pure random walk.
The collected data set consists of 1959 measurement sets (\ie laser scans) containing 2630 individual bearing measurements.

\section{Experimental Sensor Models}
\label{sec:sensor_models}
We now develop the detection, measurement, and clutter models necessary to utilize the PHD filter using the robot, sensor, and targets from the previous section.

\subsection{Detection Model}
\label{sec:detection_model}
\begin{figure}[tbp]
\centering
\subfloat[Detection model]{
    \raisebox{20mm}{\includegraphics[width=0.4\columnwidth]{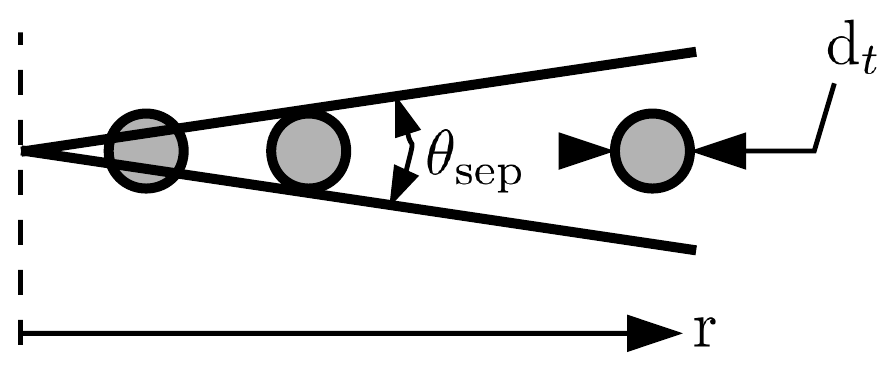}}
    \label{fig:laser_detection_model}
} \hfill
\subfloat[Clutter model]{
    \includegraphics[width=0.55\columnwidth]{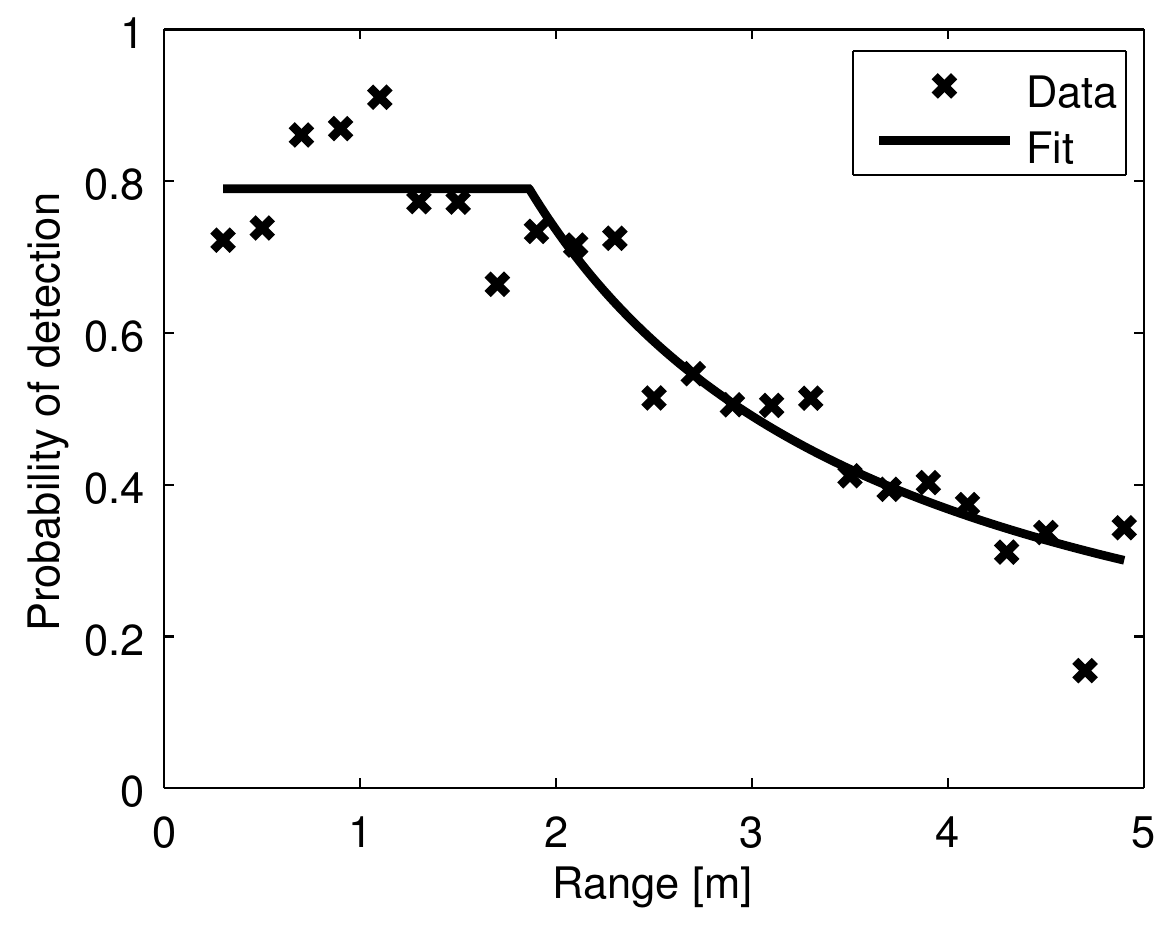}
    \label{fig:detection_model_fit}
}
\caption{(a) A pictogram of the laser detection model, where $d_t$ is the diameter of the target, $\theta_{\rm sep}$ is the angular separation between beams, and $r$ is the range. 
(b) Best fit detection model to 1588 true detections, from 1959 measurement sets, and 1007 false negative detections.}
\label{fig:detection_model}
\end{figure}

The detection model can be determined using simple geometric reasoning due to the nature of the laser scanner, as \figref{fig:laser_detection_model} shows.
Each beam in a laser scan intersects a target that is within $d_t/2$ of the beam.
The arc length between two beams at a range $r$ is $r \theta_{\rm sep}$, and the covered space is $d_t$.
Using the small angle approximation for tangent, the probability of detection is
\begin{equation}
p_d(x; q) = (1 - p_{\rm fn}) \min \left( 1, \frac{d_t}{r(x, q) \theta_{\rm sep}} \right) 
	\1{b(x, q) \in [b_{\rm min}, b_{\rm max}]} \1{r(x, q) \in [0, r_{\rm max}]}
\label{eq:DetectionModel}
\end{equation}
where $r(x, q)$ and $b(x, q)$ are the range and bearing of the target in the local sensor frame, $p_{\rm fn}$ is the probability of a false negative, and $\1{\cdot}$ is an indicator function.
The bearing is limited to fall within $[b_{\rm min}, b_{\rm max}]$ and the range to be less than some maximum value $r_{\rm max}$ (here due to the intensity threshold on the laser and the reflectivity of the targets).

To find the optimal parameter values, we take the collected data and determine which measurements originate from true objects.
A measurement is labeled as a true detection if it is within 3\,$\sigma$ of the true bearing to the target and within the field of view of the robot, given its current pose.
Since the PHD filter assumes that each target creates at most one measurement per scan, once a measurement-to-target association is made, that target is no longer fit to any other measurements in a measurement set.
If no measurement is associated to a target, the target is labeled as a false negative detection.
The collected data contains 1588 true detections and there are 1007 false negative detections.

We bin this labeled data as a function of the true range to the target in \unit[0.2]{m} increments, computing the probability of detecting a target within each range bin.
We search over $p_{\rm fn}$ (from 0 to 1 in steps of 0.005) and $d_t$ (from \unit[0]{in} to \unit[2]{in} in steps of \unit[0.01]{in}), computing the sum-of-squares error between the data and the parameterized model.
We find the effective target diameter $d_t$ since the intensity is below the cutoff threshold at extremely high angles of incidence to the reflective tape (see \figref{fig:labeled_intensity_targets}).
\figref{fig:detection_model_fit} shows the best fit model, with $p_{\rm fn} = 0.210$ and $d_t = \unit[1.28]{in}$.
These parameters are reasonable, with the effective target diameter being 78.8\% of the true target diameter.
This corresponds to a maximum angle of incidence of 52.0$^\circ$, which is orders of magnitude larger than the for glass or metal.

\subsection{Measurement Model}
The sensor returns a bearing measurement to each detected target.
We assume that bearing measurements are corrupted by zero-mean Gaussian noise with covariance $\sigma$, which is independent of the robot pose and the range and bearing to the target.
In other words,
\begin{equation}
g(z \mid x; q) = \frac{1}{\sqrt{2 \pi \sigma^2}} \exp \bigg( -\frac{\big(z - b(x, q)\big)^2}{2 \sigma^2} \bigg),
\label{eq:measurement_model}
\end{equation}
where $b(x, q)$ is the bearing of the target in the sensor frame.

During the runs, the robots occasionally experience significant errors in localization due to occlusions by transient objects, long feature-poor hallways, and displaced semi-static objects, \eg chairs.
Since no ground-truth localization data is available in the experimental environment, we fit the noise parameter $\sigma$ by searching over a range of possible values.
Note that the value of $\sigma$ affects the target-to-measurement association, and thus the detection and clutter model parameters.

\begin{figure}[tbp]
\centering
\includegraphics[width=0.65\columnwidth]{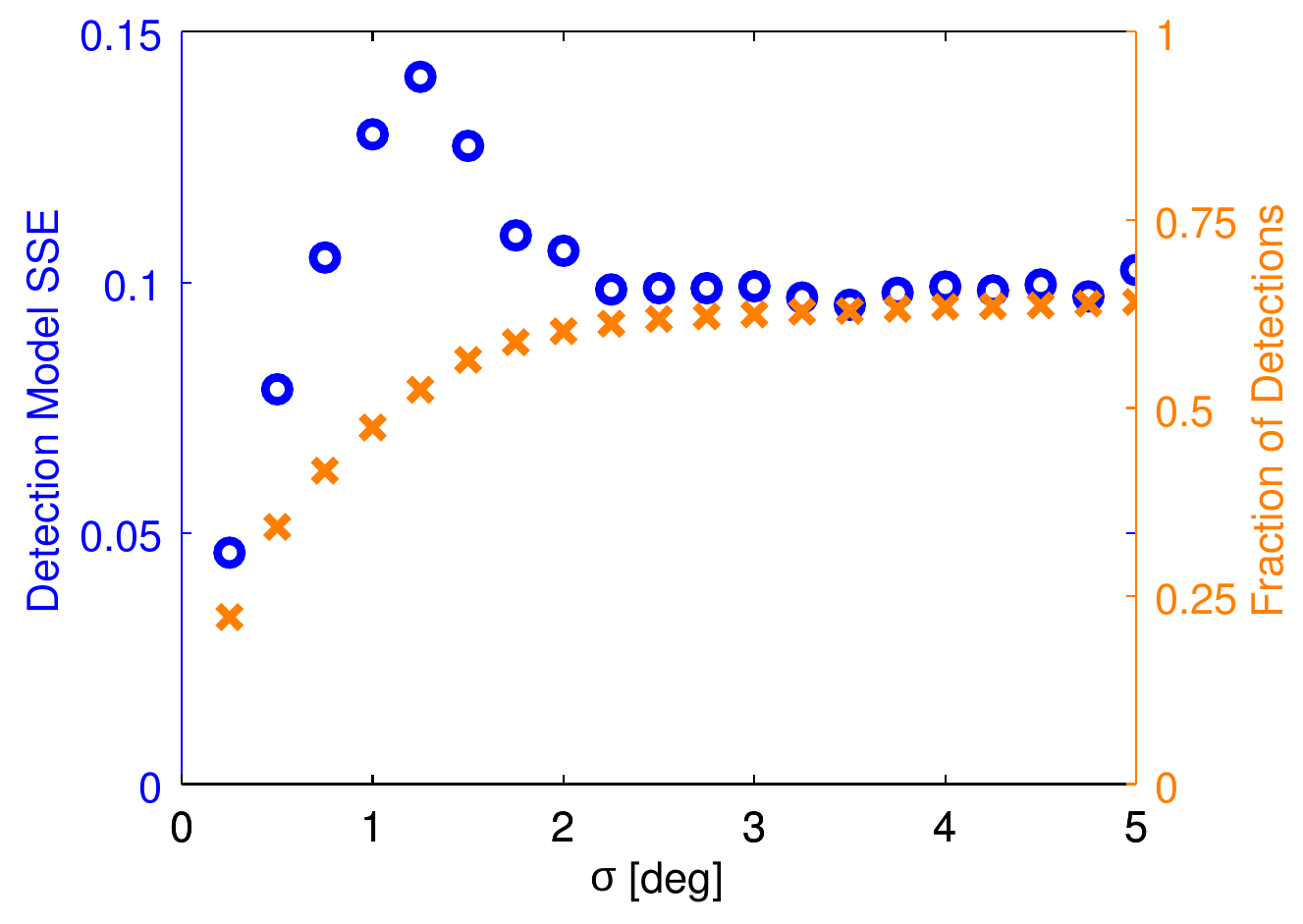}
\caption{Detection model sum of squares error (SSE) (blue circles) and the fraction of measurements that were classified as detections (orange exes) as a function of the measurement noise parameter.}
\label{fig:measurement_noise_fit}
\end{figure}

\figref{fig:measurement_noise_fit} shows that the choice of $\sigma$ parameter creates a frontier for the detection model SSE and the fraction of measurements classified as detections.
With very low values of $\sigma$, there are few inliers, \ie labeled detections, so the model fits well but is not meaningful.
The error in the fit increases until around $\sigma = 1.25^\circ$, when it begins to decrease.
From this point, the model fits increasingly well, though after a point the decrease is due to overfitting the data and there is a clear ``knee'' in the data.
This is also the point where the fraction of inliers levels out.
We select $\sigma = 2.25^\circ$ as the best fit measurement noise parameter, and use this to fit the detection and clutter models.

\subsection{Clutter Model}
\begin{figure}[tbp]
\centering
\subfloat[Clutter model]{
    \includegraphics[width=0.45\columnwidth]{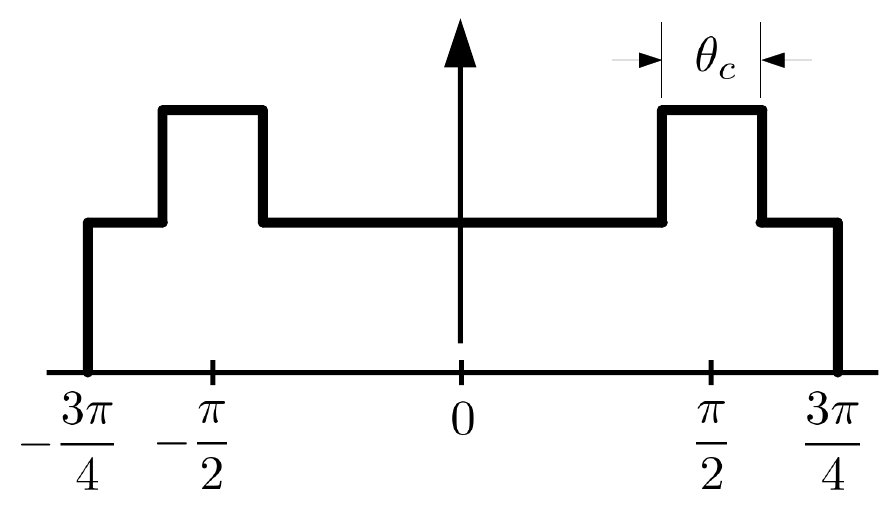}
    \label{fig:clutter_diagram}
}
\subfloat[Clutter distribution]{
	\includegraphics[width=0.52\columnwidth]{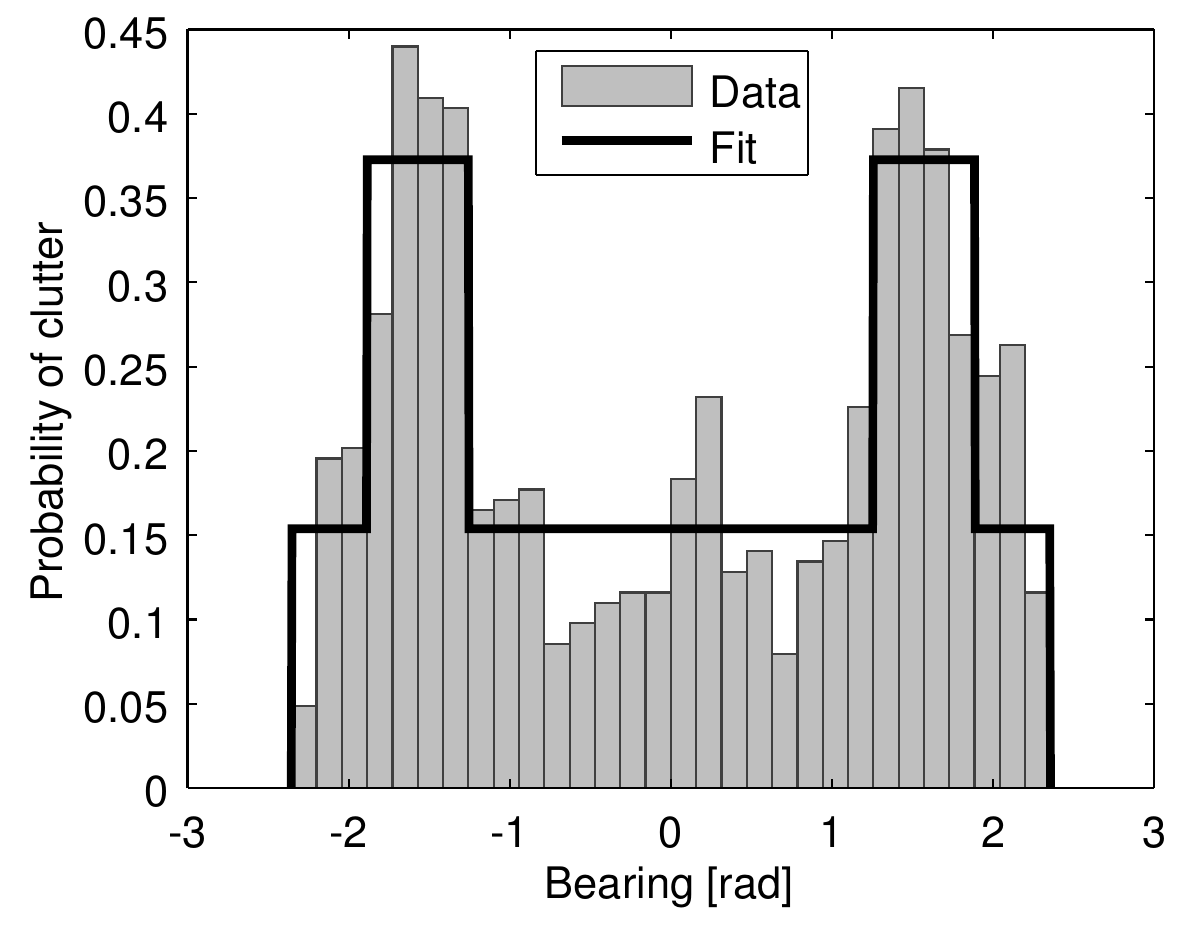}
	\label{fig:clutter_model_fit}
}
\caption{(a) A pictogram of the clutter model, where $\theta_c$ is the width of the clutter peaks centered at $\pm \frac{\pi}{2}$, and the bearing falls within the range $[-\frac{3\pi}{4}, \frac{3\pi}{4}]$.
(b) Best fit clutter probability density function, using 1010 clutter measurements from 1959 measurement sets.}
\end{figure}

As previously noted, clutter (\ie false positive) measurements arise due to reflective surfaces within the environment, such as glass and bare metal, only at low angles of incidence.
Since these materials are mostly found on walls, which are to the side of the robot when it is driving down a hallway, there will be a higher rate of clutter detections near $\pm \frac{\pi}{2}$~rad in the laser scan.
For objects such as table and chair legs there is no clear relationship between the relative pose of the object and robot, so we assume that such detections occur uniformly across the field of view of the sensor.
This leads to a clutter model of the form shown in \figref{fig:clutter_diagram}.

Let $\theta_c$ be the width of the clutter peaks centered at $\pm \frac{\pi}{2}$ and let $p_u$ be the probability that a clutter measurements was generated from a target in the uniform component of the clutter model.
The clutter model is
\begin{equation}
c(z) = \frac{p_u \mu}{b_{\rm max} - b_{\rm min}} \1{b \in [b_{\rm min}, b_{\rm max}]} + \frac{(1-p_u) \mu}{2 \theta_c} \1{\big| |b| - \pi/2 \big| \leq \theta_c/2},
\label{eq:clutter_model}
\end{equation}
where $\mu$ is the expected number of clutter measurements per scan.
The clutter cardinality, $m$, is assumed to follow a Poisson distribution with mean $\mu$ \cite{Mahler2003}.

\begin{figure}[tbp]
\centering
\includegraphics[width=0.55\columnwidth]{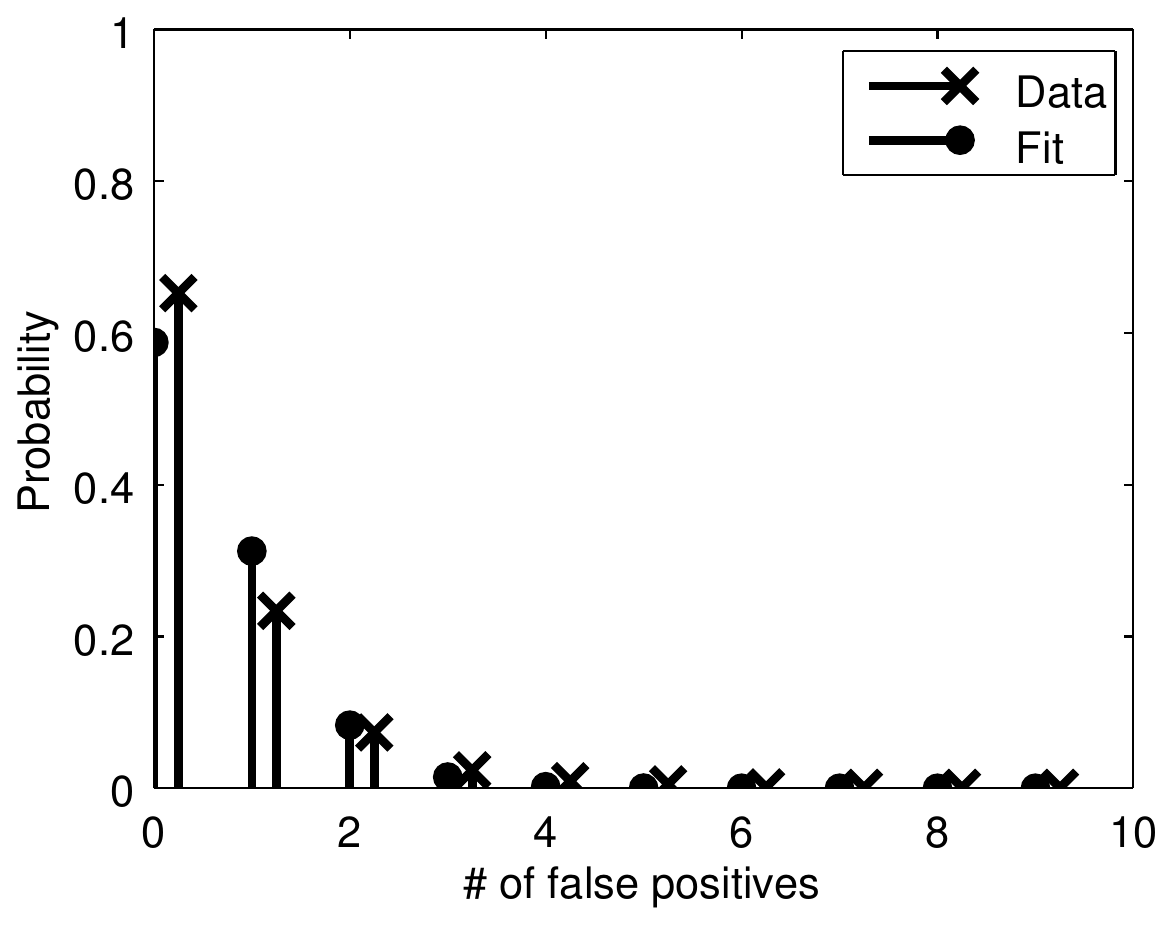}
\caption{Best fit model for the clutter cardinality, using 1010 clutter measurements from 1959 measurement sets.}
\label{fig:clutter_rate_fit}
\end{figure}

All measurements that are not associated to a target, as described in Sec.~\ref{sec:detection_model}, are considered to be clutter measurements.
We bin the bearings of these clutter measurements in $\frac{\pi}{20}$ increments to create a piecewise-constant distribution.
We perform a search over $\theta_c$ (from \unit[0]{rad} to \unit[$\frac{\pi}{2}$]{rad} in steps of \unit[$\frac{\pi}{400}$]{rad}) and $p_u$ (from 0 to 1 in steps of 0.005) to find the best fit parameters (using the sum-of-squares error) to the data, with \figref{fig:clutter_model_fit} showing the best fit model, with $\theta_c = \unit[0.200\pi]{rad}$ and $p_u = 0.725$.
The number of clutter measurements per scan is used to fit the clutter cardinality parameter $\mu$, with \figref{fig:clutter_rate_fit} showing the best fit value, $\mu = 0.5319$.

\subsection{Analysis}
All of these sensor parameters depend upon the specific robot, sensor, targets, and environment.
For example, the peaks in the clutter model near $\pm \frac{\pi}{2}$ arise due to the geometry and appearance of the environment.
In a more open setting, or with a more limited field of view sensor, we would not expect to see these peaks in the clutter distribution.
Or in an environment with more glass walls we would expect the number of clutter detections to be higher and the peaks to be more pronounced.

The detection statistics also depend highly upon our particular experimental setup.
The strip of reflective tape is only \unit[1]{in} tall and the sensor is planar, so a bump in the floor of only \unit[2.5]{mm} will cause the robot to pitch sufficiently to fail to detect a target that is \unit[1]{m} away.
Any small bumps in the linoleum flooring, particularly at transitions to carpeting, cause the robot to experience false negative detections.
Additional false negatives may occur due to occlusions from transient objects, \eg passing people and other robots, and semi-static objects such as chairs.

\section{Conclusion}
\label{sec:conclusion}
In this report we outline an experimental setup and procedure to characterize the detection, measurement, and clutter statistics of a bearing-only sensor in an indoor office environment.
While the specific values found in this work are particular to the experimental setup, we believe that by using the same parametric sensor models and by following the same experimental methodology we would be able to operate in different environments.
Getting the correct sensor statistics is particularly important when using the PHD filter, as over- or under-confidence in any of the models causes a bias in the estimate of the target cardinality.
The resulting sensor models will be used to run the PHD filter on a team of mobile robots in order to autonomously detect and localize an unknown number of objects.

\bibliographystyle{plainnat}
\bibliography{references}

\end{document}